\documentclass[11pt]{article}

\usepackage[preprint]{acl}

\usepackage{times}
\usepackage{latexsym}

\usepackage[T1]{fontenc}
\usepackage[utf8]{inputenc}

\usepackage{microtype}

\usepackage{inconsolata}

\usepackage{graphicx}
\usepackage{subcaption}
\usepackage{enumitem}

\usepackage{amsmath}
\usepackage{amsfonts}
\usepackage{array}
\usepackage{booktabs}
\usepackage{dblfloatfix}
\usepackage{float}
\usepackage{placeins}
\usepackage{url}
\usepackage[most]{tcolorbox}
\usepackage{algorithm}
\usepackage{algpseudocode}
\usepackage{geometry}
\usepackage{hyperref}
\usepackage[capitalise,nameinlink]{cleveref}
\usepackage{authblk}

\tcbset{
  findingbox/.style={
    breakable,
    colback=blue!3,
    colframe=blue!55!black,
    boxrule=0.6pt,
    arc=2pt,
    left=6pt,
    right=6pt,
    top=4pt,
    bottom=4pt,
    fonttitle=\bfseries,
  }
}
\newtcolorbox{finding}[1]{findingbox,title={#1}}
\title{Friends and Grandmothers in Silico: \\ Localizing Entity Cells in Language Models}

\author{
 \textbf{Itay Yona\textsuperscript{1}},
 \textbf{Dan Barzilay\textsuperscript{2}},
 \textbf{Michael Karasik\textsuperscript{2}},
 \textbf{Mor Geva\textsuperscript{3}}
\\
 \textsuperscript{1}Mentaleap,
 \textsuperscript{2}Independent Researcher,
 \textsuperscript{3}Tel Aviv University
\\
 \small{
   \textbf{Correspondence:} \href{mailto:itayona@gmail.com}{itay@mentaleap.ai}
 }
}

\begin{document}
\maketitle
\begin{abstract}
How do language models retrieve entity-specific facts from their parameters? We investigate this question by searching for sparse, entity-selective MLP neurons - which we call \textit{entity cells}, by analogy to the "grandmother cell" hypothesis in neuroscience - and testing whether they play a causal role in factual recall. We localize candidate entity cells by ranking MLP neurons for activation consistency across varied prompts about the same entity, applying this procedure across seven models on a curated subset of PopQA. In all models, localized neurons cluster predominantly in early layers, an empirical pattern not imposed by the architecture. Using Qwen2.5-7B base as a model organism, we find the clearest causal evidence: suppressing a localized cell selectively erases recall for its matched entity while leaving others intact, and activating a single cell is sufficient to recover correct knowledge for most entities - even when the entity is absent from the context. The same cells are recovered under aliases, acronyms, misspellings, and multilingual surface forms, and remain stable through instruction tuning, suggesting they encode canonical entity identity rather than surface token patterns. Causal signals vary across model families, pointing to architectural differences in how entity knowledge is organized. These findings offer concrete, interpretable access points for understanding, controlling, and correcting factual knowledge in language models, and draw a surprising empirical parallel to longstanding questions in neuroscience about sparse coding of concepts.
\end{abstract}

\section{Introduction}
Understanding how language models recall factual knowledge from their parameters is a core problem in mechanistic interpretability \citep{dai2022knowledgeneurons,meng2022rome,geva2023dissecting,nanda2023fact}. Many factual queries are \emph{entity-centric}: the model must resolve a named subject (e.g., \texttt{Paris} or \texttt{Barack Obama}) and then retrieve attributes about that subject. A recurring observation is that this entity processing begins early in the forward pass of the model, where token-level surface forms are transformed into semantic representations \cite{feucht2024token,kaplan2024tokens}. What is still unresolved is \emph{how and where factual access is anchored at inference time}: does the model build entity meaning gradually across many layers, or does it retrieve a compact entity representation through localized access points?

\begin{figure*}[t]
\centering
\includegraphics[width=0.65\textwidth]{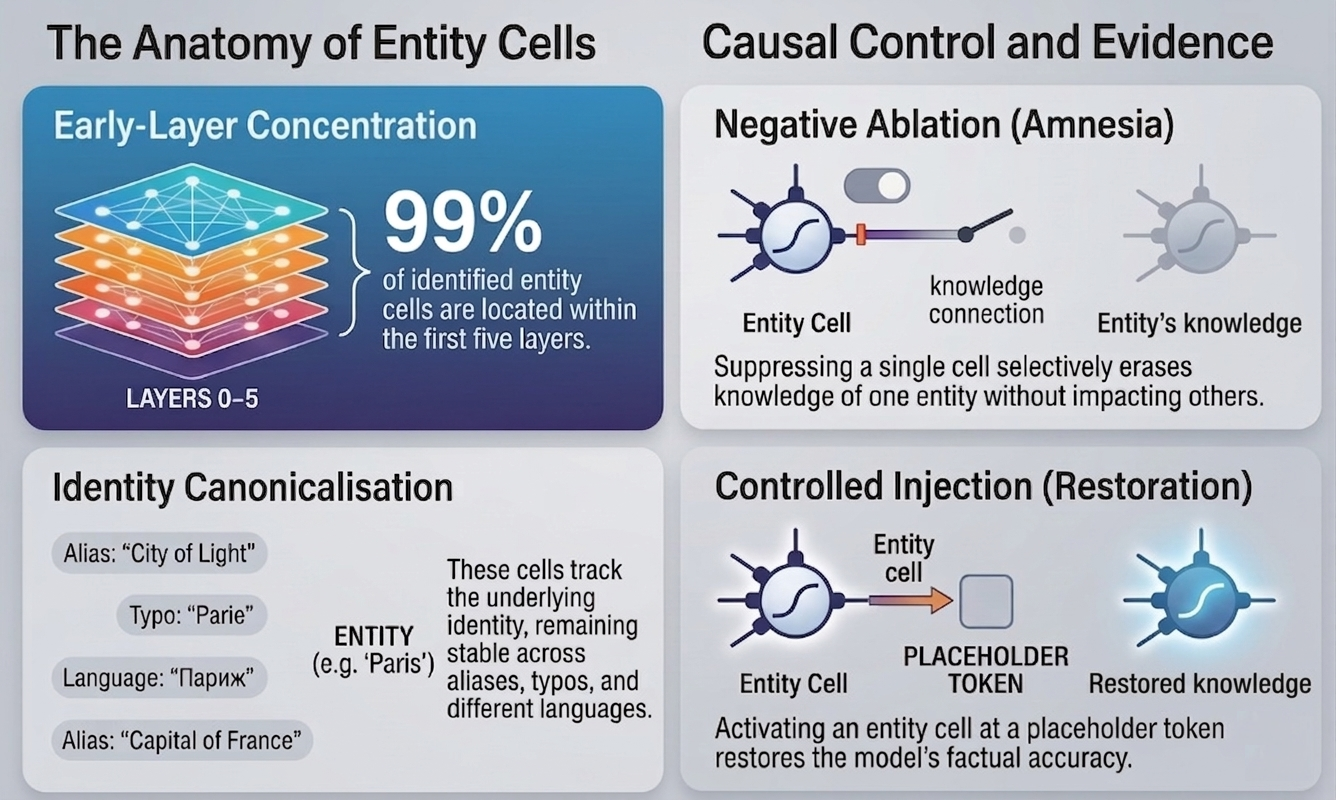}
\caption{
We identify sparse, entity-selective MLP neurons, termed \textit{entity cells}, that act as stable anchors for factual retrieval in Qwen2.5-7B. Concentrated primarily in early layers (0--5), these cells provide access to canonical identity representations that are robust to aliases, misspellings, and multilingual variants. These neurons serve as causally actionable access points: suppressing them induces entity-specific amnesia, while activating a single localized neuron is often sufficient to steer the model toward entity-consistent factual recall. Across the other six models in our suite, early-layer candidates also appear, though the causal validation is weaker.}
\label{fig:intro_overview}
\end{figure*}

By analogy to the ``grandmother cell'' hypothesis in neuroscience, we refer to sparse, entity-selective MLP neurons as \emph{entity cells}. The grandmother-cell hypothesis in neuroscience is a longstanding proposal in neuroscience, central to debates about whether individual neurons can serve as meaningful functional units in the representation of complex concepts \cite{connor2005friends,quiroga2008sparse}. In our usage, an MLP neuron is a pair of vectors within an MLP block: one vector detects a pattern in the input residual stream, and the other writes a corresponding output back to that stream \cite{geva2021kv}. Concretely, these are the matching column of $W_{\mathrm{in}}$ and row of $W_{\mathrm{out}}$.
We use \emph{entity representation} for the output written to the residual stream, or for the resulting hidden-state pattern associated
with the entity. An entity cell is therefore a neuron whose detector responds to inputs about a given entity and whose output write an
entity-consistent representation.

We investigate the existence of entity cells in LLMs using a neuron-level localization-and-intervention pipeline. Our method explicitly tests the hypothesis that an entity has a highly stable MLP neuron (identified by layer $\ell$ and neuron index $j$) at the entity mention position, across templated prompts about that entity. For example, for the entity \texttt{Donald Trump}, we use prompts such as \textit{``The origin of Donald Trump''}, \textit{``The role of Donald Trump''}, and \textit{``The location of Donald Trump''}, then record all MLP neuron activations at the final token of the entity span. We rank neurons by cross-prompt stability and take the top-ranked neuron as a candidate \emph{entity cell}.

We apply this procedure on 200 popular entities in PopQA-200, a curated subset of the PopQA dataset \cite{popqa} which serves both as the entity inventory for localization and as the source of downstream QA instances for causal evaluation. 
Across 7 models from 5 different families, we consistently observe entity-cell candidates in early layers. These models include the Qwen family (Qwen2.5-7B base, Qwen2.5-7B-Instruct, and Qwen3-8B base), OLMo-7B, Llama-3.1-8B, Mistral-7B, and OpenLLaMA-7B.

Next, given these localized candidates, we ask two questions: 
(1) does suppressing these cells impair recall about their matched entities, and (2) whether activating the cells is sufficient to restore entity knowledge in controlled settings. 
We observe the strongest trends in Qwen2.5-7B base and weaker in other model families.
In Qwen2.5, negative ablation, which scales a localized cell by a negative factor, induces entity-specific amnesia: the model becomes markedly less able to retrieve facts about the target entity, while remaining able to continue the prompt fluently and leaving control entities near baseline. Moreover, controlled injection at a placeholder token can restore access to facts about the matched entity relative to mean-entity and wrong-cell controls; for many entities, a single neuron is enough and top-$k$ adds only marginal improvements. The same localized cells also remain stable across aliases, acronyms, typos, and multilingual forms, suggesting that they provide access to entity identity across surface forms rather than a single token string.

Together, our results support a localized retrieval picture (Figure~\ref{fig:intro_overview}): for a substantial subset of entities, factual access is mediated by sparse early-layer neurons that can be causally manipulated, broadly consistent with the grandmother-cell hypothesis. We do not claim this is a full picture: reliable single-neuron effects are common but not universal, and the clearest trends are observed for popular entities.
Our work thus makes the following contributions:
\begin{enumerate}
[leftmargin=*,topsep=0pt,itemsep=3pt,parsep=0pt]
    \item We test the grandmother-cell hypothesis in language models using a stability-based method that localizes entity-sensitive neurons from templated prompts about the same entity.
    \item Applying this method across multiple models, we find the clearest and most consistent entity-cell localization and causal effects in Qwen2.5-7B base, with weaker and less consistent signals in the other tested models.
    \item We validate these localized neurons causally via negative ablation and controlled injection, including mostly single-neuron sufficiency relative to mean-entity and wrong-cell controls.
    \item We characterize key properties of localized neurons, including robustness across aliases, acronyms, misspellings, and multilingual variants.
\end{enumerate}

We release our code, prompts, and data at \url{https://github.com/1tux/in-silico/}.

\section{Related Work}
\label{sec:related}

\paragraph{Factual recall and localization}
Prior work has localized factual behavior to specific components and layers in transformers, including neuron-level interventions and causal tracing \cite{dai2022knowledgeneurons,geva2023dissecting,nanda2023fact}. Compared with Dai et al. \cite{dai2022knowledgeneurons}, which localize fact-specific neurons using paraphrases of the same fact, we localize entity-centered neurons using prompts that vary attributes of the same entity. These studies demonstrate that targeted internal changes can modulate factual outputs, but typically focus on relation-specific recall pathways. Our work is complementary: we localize \emph{entity-centered} neurons across varied relations and then test whether those neurons act as reusable access points.

\paragraph{Detokenization and entity formation}
Several studies show that early layers consolidate subword forms into coherent lexical or semantic representations \cite{elhage2022solu,feucht2024token,gurnee2023findingneuronshaystackcase,kaplan2024tokens}. We build on this line by testing whether the same localized neurons are preserved across aliases, acronyms, misspellings, and multilingual forms, linking robustness to a canonical entity representation.

\paragraph{MLP memories and sparse features}
MLP blocks have been interpreted as key-value memory mechanisms that can store and retrieve associations \cite{geva2021kv, dar2023analyzingtransformersembeddingspace}. Our findings are consistent with this view but sharpen it operationally: in many cases, a sparse neuron-level handle is enough to recover entity-consistent behavior under controlled intervention.

\paragraph{Editing and control}
Model editing methods such as ROME and MEMIT rewrite factual behavior at parameter level \cite{meng2022rome,meng2023memit}. We instead use reversible activation interventions. This isolates retrieval-time causal effects and helps separate entity access from persistent weight editing. Relative to prior work, our main contribution is a causal account of \emph{where entity-level factual access is taken from} at inference time: often from sparse, early-layer neurons that behave like compact entity access points, though not for every entity and not necessarily as the only mechanism.

We now define a localization score and the interventions used to test whether a localized neuron is merely correlational or provides causal leverage.

\section{Method}
In this section, we define the activation extraction, normalization, stability ranking, and intervention protocols used to localize and causally test entity cells.

\paragraph{Activation point}
Let $x$ be a prompt containing an entity mention and let $t(x)$ denote the entity token position. For each transformer layer $\ell$ and MLP neuron index $j$, we extract the down-projection activation of neuron $j$ at $t(x)$, denoted $a_{\ell j}(x)$. In the terminology above, the neuron is the channel-specific input-output mechanism at layer $\ell$, while $a_{\ell j}(x)$ is the scalar coefficient with which its write is applied on prompt $x$. Concretely, $a_{\ell j}(x)$ is the scalar channel value just before the MLP down-projection (\texttt{down\_proj}) at the chosen token position.

\paragraph{Normalization}
MLP activations vary widely across layers and neurons. Let $\mu_{\ell j}$ and $\sigma_{\ell j}$ denote the mean and standard deviation of $a_{\ell j}(x)$ over generic prompts $\mathcal{B}$. We standardize activations as:
\begin{equation}
z_{\ell j}(x) = \frac{a_{\ell j}(x) - \mu_{\ell j}}{\sigma_{\ell j} + \epsilon}.
\end{equation}
Unless stated otherwise we use $\epsilon=10^{-6}$.

\paragraph{Stability score and ranking}
Given a set of $K$ prompts $\{x_i\}_{i=1}^K$ that all reference the same entity, we define a stability score:
\begin{equation}
S_{\ell j} = \frac{\left(\mathbb{E}_i[z_{\ell j}(x_i)]\right)^2}{\mathrm{Std}_i[z_{\ell j}(x_i)] + \epsilon}.
\end{equation}
We rank all $(\ell,j)$ pairs by $S_{\ell j}$ and select the top neuron as the entity cell candidate for that entity. The score favors neurons that activate strongly \emph{and} consistently across entity-centered prompts.
\paragraph{Intuition} Up to $\epsilon$, $S_{\ell j}=\lvert \mathbb{E}[z]\rvert/\mathrm{CV}(z)$, where $\mathrm{CV}(z)=\mathrm{Std}(z)/\lvert \mathbb{E}[z]\rvert$ is the coefficient of variation across prompts. This makes the ranking an \emph{importance-scaled stability} criterion: high mean activation is rewarded, while high relative variability is penalized.

\paragraph{Interventions}

We employ two causal interventions on a localized cell: controlled injection and negative ablation.

\smallskip\noindent\emph{Injection.} We directly set the activation of a chosen cell at a chosen token position:
\begin{equation}
a_{\ell^\star j^\star}(x)[t(x)] \leftarrow v,
\end{equation}
with $v$ set to an entity-specific value estimated from the entity-present prompts in Finding~3. In controlled injection, this overwrite is applied on top of a mean-entity initialization, so the intervention probes directional movement on an existing entity manifold rather than de novo reconstruction from a single neuron. We use ``wrong cell'' controls by injecting a cell localized to a different entity.

\smallskip\noindent\emph{Negative ablation.} We multiply a chosen cell's activation by a scalar $\alpha$:
\begin{equation}
a_{\ell^\star j^\star}(x) \leftarrow \alpha \, a_{\ell^\star j^\star}(x),
\end{equation}
including $\alpha<0$, which flips the sign of the activation. In our implementation we apply this scaling across token positions; the effect is driven primarily by positions where the cell would otherwise activate.

\paragraph{Evaluation metrics}
Several experiments use next-token probabilities. Given a set of answer aliases $\mathcal{A}$, we define the answer score as the probability of the first token of the best-matching alias:
\begin{equation}
p_{\mathrm{ans}}(x) = \max_{a \in \mathcal{A}} p\big(\mathrm{tok}_1(a)\,|\,x\big).
\end{equation}
We use this first-token score primarily as a filtering signal when defining \emph{trustworthy} localized cells for controlled injection. In particular, the trust filter compares the target entity against no-injection and wrong-cell controls using a normalized score $\mathrm{RelProb}$, defined as the mean $p_{\mathrm{ans}}$ under a condition divided by the mean under the corresponding entity-present prompt (so 1.0 indicates parity). For injection experiments we  report \emph{pass@$k$}: whether any correct answer first-token ID appears in the top-$k$ next-token distribution (we use $k{=}5$ unless stated otherwise). This is computed directly from the next-token logits (top-$k$ membership), without sampling.
For entity-specific amnesia tests (main Finding~2) we define a normalized score based on log-probabilities, anchored by an unknown-entity baseline computed by swapping the entity name for a small set of unseen names and averaging the resulting answer log-probabilities.

\section{Experimental Setup}
\label{sec:setup}
\paragraph{Models}
We run localization and causal checks on Qwen2.5-7B base and Qwen2.5-7B-Instruct \cite{qwen25}, Qwen3-8B base \cite{qwen3}, OLMo-7B-0724-hf \cite{olmo7b}, Llama-3.1-8B-Instruct \cite{llama31}, Mistral-7B-v0.3 \cite{mistral7b}, and OpenLLaMA-7B \cite{openllama}. Section~\ref{sec:results} focuses on Qwen2.5-7B base, with cross-model comparisons reported in Appendices~\ref{app:qwen3} and~\ref{app:cross_model}. Unless stated otherwise, we run inference in half precision with automatic device mapping.

\paragraph{Data}
We use PopQA \cite{popqa}, an entity-centric QA dataset derived from Wikidata with subject entities and answer aliases.

We build a curated set of $N=200$ popular entities by seeding countries, cities, and widely known people, then filling from PopQA by popularity with a minimum of two available questions per entity. We denote this subset as PopQA-200. PopQA-200 serves as the entity inventory for localization and as the source of downstream QA examples for causal evaluation. For PopQA-based causal checks we use $K=2$ questions per entity.

When a question does not contain a recoverable entity span after tokenization, we skip it for position-dependent analyses.

\paragraph{Prompting}
For localization, we use templated prompts about each entity; examples are listed in Appendix~\ref{app:prompts}. The entity token position is defined as the final token in the tokenized entity span.

For PopQA-based evaluation, we format each question as:
\begin{quote}
\texttt{``Question: <question>\textbackslash{}nAnswer:''}
\end{quote}
For generic probing prompts (used in baselines and controlled interventions), we use cloze-style completions of the form \texttt{``Fact: ...''} (Appendix~\ref{app:prompts}).

For each localization or evaluation prompt, we locate the \emph{entity token position} as the final token in the tokenized subject-entity span. Interventions that target the entity position act at this index; cloze-style prompts define an analogous entity position at the dummy placeholder token \texttt{X}.

\paragraph{Normalization statistics}
To normalize activations across layers and neurons, we compute baseline statistics $(\mu_{\ell j}, \sigma_{\ell j})$ using 399 generic prompts (Appendix~\ref{app:prompts}), extracting activations at the final token position of each prompt. Baseline prompts are deliberately \emph{not} entity specific.

\paragraph{Implementation and compute}
We trace activations and apply in-graph interventions using NNsight \cite{nnsight}, a tracing library that exposes intermediate activations at inference time. All experiments were executed on a single GPU (NVIDIA A100).

\begin{figure}[t]
\centering
\includegraphics[width=\columnwidth]{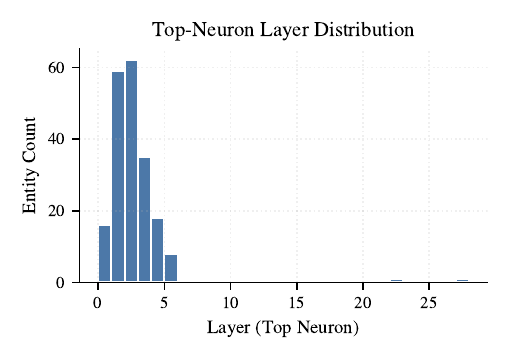}
\caption{Layer of the top localized cell for each PopQA-200 entity in Qwen2.5-7B base (n=200). Similar early-layer concentration is observed across other tested models; see Appendix~\ref{app:cross_model}.}
\label{fig:layerhist}
\end{figure}

\section{Results}
\label{sec:results}
We report four results that progressively strengthen evidence from correlational localization to causal leverage. All analyses were run on the full suite of seven models described in Section~\ref{sec:setup}. Early-layer concentration (Finding~1) is a recurring pattern across model families; similar trends appear in Qwen2.5-7B-Instruct, Qwen3-8B, and to a lesser extent in the other models tested (Appendices~\ref{app:qwen3}--\ref{app:cross_model}). The main text focuses on Qwen2.5-7B base, which yields the strongest and most consistent causal evidence across all four findings; cross-model results are summarized in the appendix. We first map where sparse entity cells appear (Finding~1), then test whether suppressing and activating a localized neuron affects entity-specific recall (Findings~2--3). Finally, we use surface-form perturbations as an interpretive check on what information these cells provide access to (Finding~4). Unless noted otherwise, localization uses the PopQA-200 entity set together with the templated prompts described in Section~\ref{sec:setup}. The full 200-entity cell map with trustworthiness flags is provided in Appendix~\ref{app:entity_neurons}.

\subsection{Localizing Entity Cells}
\begin{finding}{Finding 1}
Entity cells concentrate in early layers (0--5), without being imposed by the architecture
\end{finding}
For each entity, we rank all MLP neurons (indexed by layer $\ell$ and neuron $j$) by stability at the mention position across $K$ prompts and record the layer of the top-ranked cell. Localization is strongly non-uniform (Figure~\ref{fig:layerhist}): 99.0\% of entities peak in layers 0--5, and only 1.0\% peak in layers 22 or 27. Since ranking is over all 28 layers, this depth profile is empirical rather than enforced, and is consistent with early canonicalization features that help form an entity identity representation.

Early-layer concentration is suggestive, but does not establish that a localized neuron matters for factual extraction. We next test whether suppressing a candidate entity cell selectively impairs recall about that entity.

\subsection{Causal Necessity}
\begin{figure}[t]
\centering
\includegraphics[width=\columnwidth]{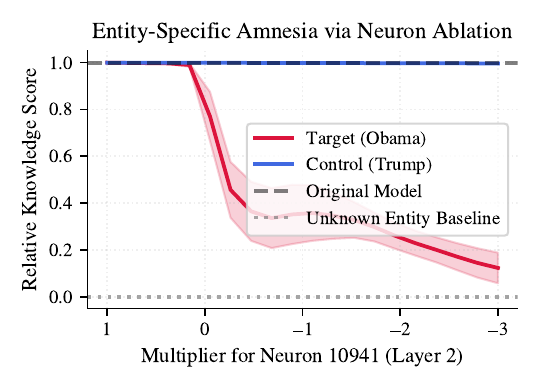}
\caption{Entity-specific amnesia under negative ablation for the localized Obama cell (L2-N10941). Target (Obama) recall drops substantially as $\alpha$ decreases, while control (Trump) remains near baseline.}
\label{fig:f6}
\end{figure}

\begin{finding}{Finding 2}
Negatively ablating a localized cell selectively suppresses recall for the target entity while leaving control entities near baseline. At scale, 131/200 localized cells show this entity-specific effect
\end{finding}
We apply negative ablation (a signed multiplier) to localized cells and measure entity-specific recall while checking for pathological collapse. In a case study, target retention drops from 1.0 to 0.123 at $\alpha=-3$, while a control entity (Trump) stays near baseline (1.0 to 0.996; Figure~\ref{fig:f6}). This behavior is consistent with the localized neuron being part of the access path to many entity-linked facts: the model still processes the prompt, but loses the identity representation needed for reliable recall. We then run the same criterion at scale and use it to define a \emph{trustworthy} localized cell: a neuron whose suppression produces substantial entity-specific loss without destabilizing the model. Under these checks, 131/200 localized cells are marked trustworthy and define the subset used for controlled injection.

Having used negative ablation to establish necessity and to filter trustworthy neurons, we next test whether activating a single cell is sufficient to steer output in a controlled placeholder setting.

\subsection{Causal Sufficiency}
\begin{figure}[t]
\centering
\includegraphics[width=\columnwidth]{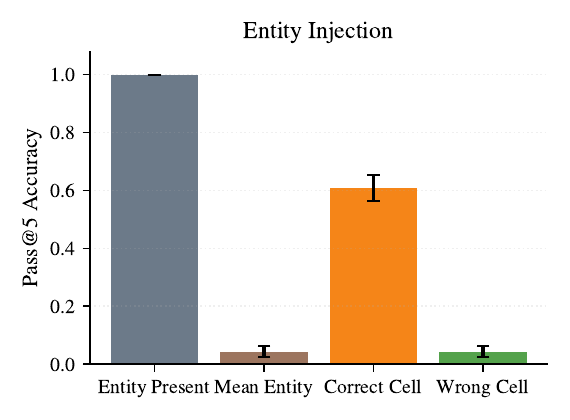}
\caption{Controlled injection at the placeholder token \texttt{X}, evaluated on instances where the entity-present prompt is already correct under pass@5 (109 examples). Mean-entity initialization and wrong-cell injection are control conditions; correct-cell injection shows the expected directional gain.}
\label{fig:f4}
\end{figure}

\begin{finding}{Finding 3}
Correct-cell injection recovers entity-specific recall (63.3\% pass@5) against near-zero controls, with a single cell being sufficient
\end{finding}
On the trustworthy subset from Finding~2 (Appendix~\ref{app:entity_neurons}), entity-present pass@5 is 109/262 (41.6\%) across evaluated question instances. To isolate intervention effects from base-model misses, we report injection on the 109 instances where the entity-present prompt is already correct under pass@5. We replace the entity mention with \texttt{X} and intervene at the placeholder token. Mean-entity initialization and wrong-cell injection are used as controls. On this known-answer subset, pass@5 is 1.8\% for mean-entity control, 63.3\% for correct-cell injection, and 1.8\% for wrong-cell injection (Figure~\ref{fig:f4}). Single-cell injection remains largely sufficient: 41/79 entities pass with top-1 versus 42/79 with top-$k$; only one entity requires multi-cell injection. We select $\alpha$ per entity from a small grid, which improves sensitivity but can be optimistic relative to a fixed-$\alpha$ protocol.
For {\ttfamily Question: Who is the spouse of X?\textbackslash nAnswer:}, we set the hidden vector at \texttt{X} to the mean-entity vector, then activate the Obama cell at that same token position.

The causal results above establish necessity (ablation) and sufficiency (injection) for entity-specific recall in this protocol. We now ask what information the localized neuron provides access to, by testing stablity under surface-form variations.

\subsection{Surface-Form Robustness}
\begin{figure}[t]
\centering
\includegraphics[width=\columnwidth]{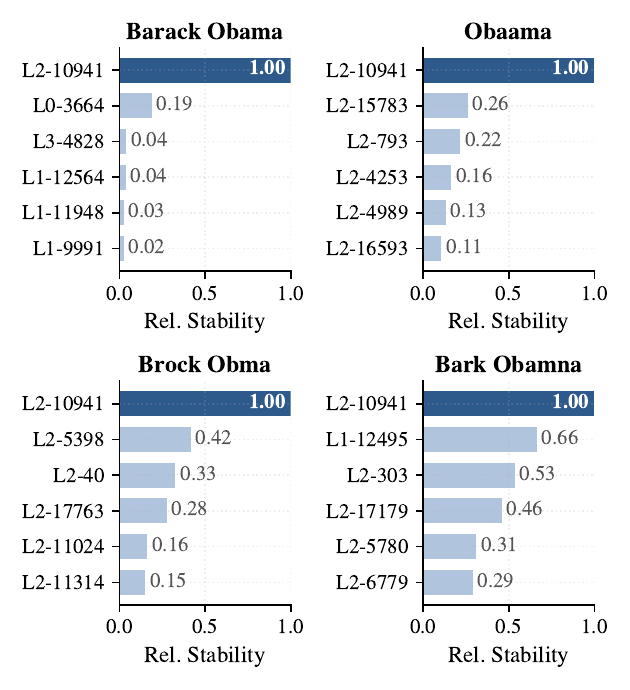}
\caption{Variant robustness for ``Barack Obama'': most spelling and phrasing perturbations keep the same localized cell (L2-N10941).}
\label{fig:f3}
\end{figure}

\begin{figure}[t]
\centering
\includegraphics[width=\columnwidth]{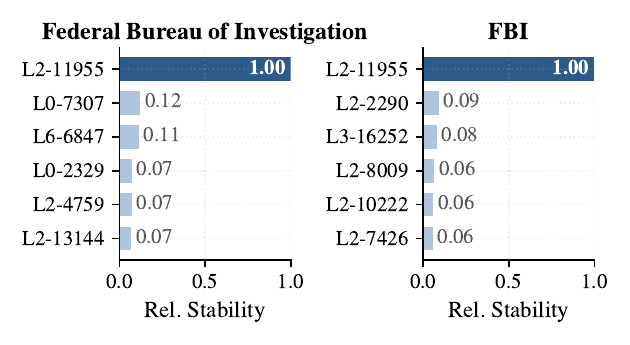}
\caption{Acronym robustness (FBI): variants localize to the same top-ranked cell (L2-N11955).}
\label{fig:f3_acronym}
\end{figure}

\begin{figure}[t]
\centering
\includegraphics[width=\columnwidth]{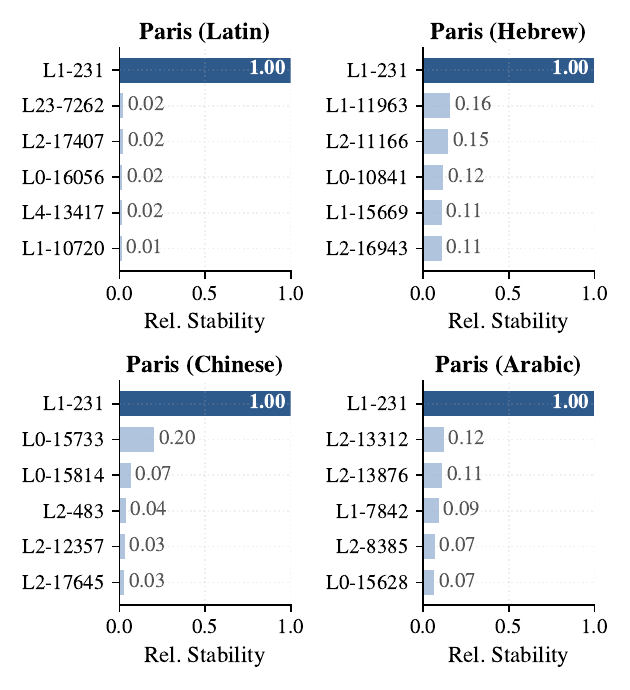}
\caption{Multilingual robustness (Paris): variants localize to the same top-ranked cell (L1-N231).}
\label{fig:f3_multilingual}
\end{figure}

\begin{finding}{Finding 4}
The same cell is recovered across spelling variants, acronyms, and multilingual forms, suggesting access to a canonical identity representation
\end{finding}
We re-run localization on the same prompt templates while perturbing the entity string. We test spelling/phrasing variants (Barack Obama), acronym variants (FBI), and multilingual variants (Paris). Most spelling and phrasing variants of ``Barack Obama'' preserve the same top cell (L2-N10941) in Figure~\ref{fig:f3}. We observe similar robustness for acronym and multilingual surface forms (Figures~\ref{fig:f3_acronym} and~\ref{fig:f3_multilingual}), consistent with an identity-canonicalization role rather than dependence on a single token sequence.

\section{Discussion}
The combined evidence supports a canonicalization-and-control view of entity cells in Qwen2.5-7B base. Negative ablation induces entity-specific amnesia and provides a practical trust filter at scale (131/200), indicating that localized neurons are functionally necessary for entity-specific recall in this protocol. On the known-answer subset of Finding~3 (109/262 instances), controlled injection is strongly directional and mostly single-cell sufficient (41/79 with top-1 vs.\ 42/79 with top-$k$), providing a complementary sufficiency test. Robustness to typos, acronyms, and multilingual forms suggests that these neurons provide access to identity-level information rather than a single token string, and the early-layer concentration is consistent with a role in forming an entity identity representation used for downstream factual extraction. Taken together, the cells behave like a latent \emph{entity vocabulary}: sparse anchor neurons that point computation toward an entity-consistent state and thereby gate access to distributed factual circuits.

The appendix broadens the scope of the main result by testing the same pipeline across multiple models. The strongest extension is post-training robustness: Qwen2.5-7B-Instruct preserves nearly the same entity-cell map as the base model. Qwen3-8B also exhibits sparse early-layer entity cells under the same localization procedure, although the causal evidence is weaker. Across other model families, candidate cells can often still be localized, but trustworthy causal effects and form robustness are much less consistent. Taken together, this suggests that entity cells are a reproducible but model-dependent phenomenon.

\section{Limitations and Scope}
This study focuses on one dataset (PopQA), with the strongest and most complete evidence in Qwen2.5-7B base. This potentially could be explained by pretraining-data composition: Qwen documentation reports strongest capability in English and Chinese, with broader multilingual
performance depending on available data coverage
\cite{qwen_intro,qwen2_blog}. If so, mechanism visibility may be data-distribution-dependent, so generalization to other model families
should be treated as an empirical question and tested with like-for-like replications.

Our localization score is intentionally sparse: it ranks individual neurons first, and in Finding~3 top-$k$ variants provided only marginal
gains over top-1. This design prioritizes interpretability but may still miss distributed or multi-cell codes \cite{shafran2025decomposing}. We use $K{=}2$ prompts per entity for localization and causal checks, which can introduce per-entity
instability.

Our metrics are mostly first-token based, which can understate
multi-token factual competence and can reflect lexical priming effects.
In Finding~3, $\alpha$ is selected per entity from a sweep; this
improves sensitivity but can introduce optimistic bias, and a fixed-$\alpha$
protocol is an important next step. Our injection and ablation
experiments are still narrow in relation coverage. We also include an
exploratory \emph{factual modification} procedure via latent steering:
optimizing a small perturbation injected at an entity-associated
activation site to rewrite a specific relation while preserving
unrelated facts; Appendix~\ref{app:latent_steering} provides a concrete
template (Algorithm~\ref{alg:steer}) and prompt set.

\section{Conclusion}
We test the ``grandmother cell'' hypothesis from neuroscience across multiple language model families.
In Qwen2.5-7B base, we find sparse, stable, and causally actionable entity cells: they concentrate in early layers, negative ablation induces entity-specific amnesia, and controlled injection is mostly single-neuron sufficient on the known-answer subset. Robustness to surface-form variation supports the view that these cells provide access to identity-level information. The appendix shows that the phenomenon extends with different strength across additional models. The clearest extension is post-training robustness in Qwen2.5-7B-Instruct, while Qwen3-8B also exhibits sparse early-layer entity cells under the same pipeline. Across other model families, the signal is weaker and less consistent, suggesting that entity cells are reproducible but model-dependent access points for factual retrieval.

\section*{Acknowledgements}
We thank Katja Filippova for invaluable insights and perspective on this work.

\bibliography{refs}

@inproceedings{dai2022knowledgeneurons,
  title     = {Knowledge Neurons in Pretrained Transformers},
  author    = {Dai, Damai and Dong, Li and Hao, Yaru and Sui, Zhifang and Wei, Furu},
  booktitle = {Proceedings of the 60th Annual Meeting of the Association for Computational Linguistics (ACL)},
  year      = {2022},
  doi       = {10.18653/v1/2022.acl-long.581},
  url       = {https://arxiv.org/abs/2104.08696}
}

@inproceedings{meng2022rome,
  title     = {Locating and Editing Factual Associations in {GPT}},
  author    = {Meng, Kevin and Bau, David and Andonian, Alex and Belinkov, Yonatan},
  booktitle = {Advances in Neural Information Processing Systems (NeurIPS)},
  year      = {2022},
  url       = {https://arxiv.org/abs/2202.05262}
}

@inproceedings{meng2023memit,
  title     = {Mass-Editing Memory in a Transformer},
  author    = {Meng, Kevin and Sharma, Arnab Sen and Andonian, Alex and Belinkov, Yonatan and Bau, David},
  booktitle = {International Conference on Learning Representations (ICLR)},
  year      = {2023},
  url       = {https://arxiv.org/abs/2210.07229}
}

@inproceedings{geva2021kv,
  title     = {Transformer Feed-Forward Layers Are Key-Value Memories},
  author    = {Geva, Mor and Schuster, Roei and Berant, Jonathan and Levy, Omer},
  booktitle = {Proceedings of the 2021 Conference on Empirical Methods in Natural Language Processing (EMNLP)},
  year      = {2021},
  doi       = {10.18653/v1/2021.emnlp-main.446},
  url       = {https://arxiv.org/abs/2012.14913}
}

@misc{qwen25,
      title={Qwen2.5 Technical Report}, 
      author={An Yang and Baosong Yang and Beichen Zhang and Binyuan Hui and Bo Zheng and Bowen Yu and Chengyuan Li and Dayiheng Liu and Fei Huang and Haoran Wei and Huan Lin and Jian Yang and Jianhong Tu and Jianwei Zhang and Jianxin Yang and Jiaxi Yang and Jingren Zhou and Junyang Lin and Kai Dang and Keming Lu and Keqin Bao and Kexin Yang and Le Yu and Mei Li and Mingfeng Xue and Pei Zhang and Qin Zhu and Rui Men and Runji Lin and Tianhao Li and Tianyi Tang and Tingyu Xia and Xingzhang Ren and Xuancheng Ren and Yang Fan and Yang Su and Yichang Zhang and Yu Wan and Yuqiong Liu and Zeyu Cui and Zhenru Zhang and Zihan Qiu},
      year={2025},
      eprint={2412.15115},
      archivePrefix={arXiv},
      primaryClass={cs.CL},
      url={https://arxiv.org/abs/2412.15115}, 
}

@misc{qwen3,
      title={Qwen3 Technical Report}, 
      author={An Yang and Anfeng Li and Baosong Yang and Beichen Zhang and Binyuan Hui and Bo Zheng and Bowen Yu and Chang Gao and Chengen Huang and Chenxu Lv and Chujie Zheng and Dayiheng Liu and Fan Zhou and Fei Huang and Feng Hu and Hao Ge and Haoran Wei and Huan Lin and Jialong Tang and Jian Yang and Jianhong Tu and Jianwei Zhang and Jianxin Yang and Jiaxi Yang and Jing Zhou and Jingren Zhou and Junyang Lin and Kai Dang and Keqin Bao and Kexin Yang and Le Yu and Lianghao Deng and Mei Li and Mingfeng Xue and Mingze Li and Pei Zhang and Peng Wang and Qin Zhu and Rui Men and Ruize Gao and Shixuan Liu and Shuang Luo and Tianhao Li and Tianyi Tang and Wenbiao Yin and Xingzhang Ren and Xinyu Wang and Xinyu Zhang and Xuancheng Ren and Yang Fan and Yang Su and Yichang Zhang and Yinger Zhang and Yu Wan and Yuqiong Liu and Zekun Wang and Zeyu Cui and Zhenru Zhang and Zhipeng Zhou and Zihan Qiu},
      year={2025},
      eprint={2505.09388},
      archivePrefix={arXiv},
      primaryClass={cs.CL},
      url={https://arxiv.org/abs/2505.09388}, 
}

@inproceedings{olmo7b,
    title = "{OLM}o: Accelerating the Science of Language Models",
    author = "Groeneveld, Dirk  and
      Beltagy, Iz  and
      Walsh, Evan  and
      Bhagia, Akshita  and
      Kinney, Rodney  and
      Tafjord, Oyvind  and
      Jha, Ananya  and
      Ivison, Hamish  and
      Magnusson, Ian  and
      Wang, Yizhong  and
      Arora, Shane  and
      Atkinson, David  and
      Authur, Russell  and
      Chandu, Khyathi  and
      Cohan, Arman  and
      Dumas, Jennifer  and
      Elazar, Yanai  and
      Gu, Yuling  and
      Hessel, Jack  and
      Khot, Tushar  and
      Merrill, William  and
      Morrison, Jacob  and
      Muennighoff, Niklas  and
      Naik, Aakanksha  and
      Nam, Crystal  and
      Peters, Matthew  and
      Pyatkin, Valentina  and
      Ravichander, Abhilasha  and
      Schwenk, Dustin  and
      Shah, Saurabh  and
      Smith, William  and
      Strubell, Emma  and
      Subramani, Nishant  and
      Wortsman, Mitchell  and
      Dasigi, Pradeep  and
      Lambert, Nathan  and
      Richardson, Kyle  and
      Zettlemoyer, Luke  and
      Dodge, Jesse  and
      Lo, Kyle  and
      Soldaini, Luca  and
      Smith, Noah  and
      Hajishirzi, Hannaneh",
    editor = "Ku, Lun-Wei  and
      Martins, Andre  and
      Srikumar, Vivek",
    booktitle = "Proceedings of the 62nd Annual Meeting of the Association for Computational Linguistics (Volume 1: Long Papers)",
    month = aug,
    year = "2024",
    address = "Bangkok, Thailand",
    publisher = "Association for Computational Linguistics",
    url = "https://aclanthology.org/2024.acl-long.841/",
    doi = "10.18653/v1/2024.acl-long.841",
    pages = "15789--15809",
}

@article{llama31,
  title        = {The {Llama~3} Herd of Models},
  author       = {Aaron Grattafiori and Abhimanyu Dubey and Abhinav Jauhri and
                  Abhinav Pandey and Abhishek Kadian and Ahmad Al-Dahle and others},
  journal      = {arXiv preprint arXiv:2407.21783},
  year         = {2024},
  url          = {https://arxiv.org/abs/2407.21783}
}

@misc{mistral7b,
      title={Mistral 7B}, 
      author={Albert Q. Jiang and Alexandre Sablayrolles and Arthur Mensch and Chris Bamford and Devendra Singh Chaplot and Diego de las Casas and Florian Bressand and Gianna Lengyel and Guillaume Lample and Lucile Saulnier and Lélio Renard Lavaud and Marie-Anne Lachaux and Pierre Stock and Teven Le Scao and Thibaut Lavril and Thomas Wang and Timothée Lacroix and William El Sayed},
      year={2023},
      eprint={2310.06825},
      archivePrefix={arXiv},
      primaryClass={cs.CL},
      url={https://arxiv.org/abs/2310.06825}, 
}

@misc{openllama,
  author = {Geng, Xinyang and Liu, Hao},
  title = {OpenLLaMA: An Open Reproduction of LLaMA},
  month = May,
  year = 2023,
  url = {https://github.com/openlm-research/open_llama}
}

@misc{popqa ,
  title={When Not to Trust Language Models: Investigating Effectiveness and Limitations of Parametric and Non-Parametric Memories },
  author={ Mallen, Alex and Asai,Akari and  Zhong, Victor and Das, Rajarshi and Hajishirzi, Hannaneh and Khashabi, Daniel},
  journal={ arXiv preprint },
  year={ 2022 }
}

@article{connor2005friends,
  title={Friends and grandmothers},
  author={Connor, Charles E},
  journal={Nature},
  volume={435},
  number={7045},
  pages={1036--1037},
  year={2005},
  publisher={Nature Publishing Group UK London}
}

@article{quiroga2008sparse,
  title={Sparse but not 'grandmother-cell' coding in the medial temporal lobe},
  author={Quiroga, R. Quian and Kreiman, Gabriel and Koch, Christof and Fried, Itzhak},
  journal={Trends in Cognitive Sciences},
  volume={12},
  number={3},
  pages={87--91},
  year={2008},
  publisher={Elsevier}
}

@article{geva2023dissecting,
  title={Dissecting recall of factual associations in auto-regressive language models},
  author={Geva, Mor and Bastings, Jasmijn and Filippova, Katja and Globerson, Amir},
  journal={arXiv preprint arXiv:2304.14767},
  year={2023},
  url={https://arxiv.org/abs/2304.14767}
}

@inproceedings{nanda2023fact,
  title={Fact finding: Attempting to reverse-engineer factual recall on the neuron level},
  author={Nanda, Neel and Rajamanoharan, Senthooran and Kramar, Janos and Shah, Rohin},
  booktitle={Alignment Forum},
  year={2023}
}

@misc{gurnee2023findingneuronshaystackcase,
  title={Finding Neurons in a Haystack: Case Studies with Sparse Probing},
  author={Gurnee, Wes and Nanda, Neel and Pauly, Matthew and Harvey, Katherine and Troitskii, Dmitrii and Bertsimas, Dimitris},
  year={2023},
  eprint={2305.01610},
  archivePrefix={arXiv},
  primaryClass={cs.LG},
  url={https://arxiv.org/abs/2305.01610}
}

@article{feucht2024token,
  title={Token Erasure as a Footprint of Implicit Vocabulary Items in LLMs},
  author={Feucht, Sheridan and Atkinson, David and Wallace, Byron and Bau, David},
  journal={arXiv preprint arXiv:2406.20086},
  year={2024},
  url={https://arxiv.org/abs/2406.20086}
}

@article{kaplan2024tokens,
  title={From Tokens to Words: On the Inner Lexicon of LLMs},
  author={Kaplan, Guy and Oren, Matanel and Reif, Yuval and Schwartz, Roy},
  journal={arXiv preprint arXiv:2410.05864},
  year={2024},
  url={https://arxiv.org/abs/2410.05864}
}

@misc{dar2023analyzingtransformersembeddingspace,
  title={Analyzing Transformers in Embedding Space},
  author={Dar, Guy and Geva, Mor and Gupta, Ankit and Berant, Jonathan},
  year={2023},
  eprint={2209.02535},
  archivePrefix={arXiv},
  primaryClass={cs.CL},
  url={https://arxiv.org/abs/2209.02535}
}

@article{elhage2022solu,
  title={Softmax Linear Units},
  author={Elhage, Nelson and Hume, Tristan and Olsson, Catherine and Nanda, Neel and Henighan, Tom and Johnston, Scott and ElShowk, Sheer and Joseph, Nicholas and DasSarma, Nova and Mann, Ben and Hernandez, Danny and Askell, Amanda and Ndousse, Kamal and Jones, Andy and Drain, Dawn and Chen, Anna and Bai, Yuntao and Ganguli, Deep and Lovitt, Liane and Hatfield-Dodds, Zac and Kernion, Jackson and Conerly, Tom and Kravec, Shauna and Fort, Stanislav and Kadavath, Saurav and Jacobson, Josh and Tran-Johnson, Eli and Kaplan, Jared and Clark, Jack and Brown, Tom and McCandlish, Sam and Amodei, Dario and Olah, Christopher},
  year={2022},
  journal={Transformer Circuits Thread},
  url={https://transformer-circuits.pub/2022/solu/index.html}
}

@misc{nnsight,
  title={NNsight: Library for Interpreting Language Models},
  author={{NDIF Team}},
  year={2024},
  howpublished={\url{https://github.com/ndif-team/nnsight}},
  note={Accessed: 2026-02-27}
}

@misc{qwen_intro,
  title        = {Introducing Qwen},
  author       = {{Qwen Team}},
  howpublished = {QwenLM Blog},
  year         = {2023},
  note         = {Accessed 2026-02-28},
  url          = {https://qwenlm.github.io/blog/qwen/}
}

@misc{qwen2_blog,
  title        = {Qwen2: Better than Ever},
  author       = {{Qwen Team}},
  howpublished = {QwenLM Blog},
  year         = {2024},
  note         = {Accessed 2026-02-28},
  url          = {https://qwenlm.github.io/blog/qwen2/}
}

@article{shafran2025decomposing,
  title={Decomposing mlp activations into interpretable features via semi-nonnegative matrix factorization},
  author={Shafran, Or and Geiger, Atticus and Geva, Mor},
  journal={arXiv preprint arXiv:2506.10920},
  year={2025}
}

\clearpage
\appendix
\section{Prompt Templates and Hyperparameters}
\label{app:prompts}
Table~\ref{tab:prompts} lists the prompt templates used across experiments. Baseline prompts consist of 399 generic cloze-style statements (e.g., ``The Eiffel Tower is located in''), used only to estimate $(\mu,\sigma)$ for normalization.

\begin{table}[!ht]
\centering
\scriptsize
\begingroup
\setlength{\tabcolsep}{2pt}
\renewcommand{\arraystretch}{1.08}
\begin{tabular}{@{}
  >{\raggedright\arraybackslash}p{0.27\columnwidth}
  >{\raggedright\arraybackslash}p{0.67\columnwidth}
@{}}
\toprule
\textbf{Use} & \textbf{Template} \\
\midrule
PopQA QA wrapper & {\ttfamily Question: <q>\allowbreak\textbackslash{}nAnswer:} \\
Generic baselines & {\ttfamily <statement fragment>} \\
Localization probes & {\ttfamily The <attribute> of <entity>} \\
Injection/ablation & {\ttfamily Fact: the <relation> of <entity>:} \\
Factual modification & {\ttfamily The spouse of <entity> is named} \\
\bottomrule
\end{tabular}
\endgroup
\caption{Prompt templates used in this work.}
\label{tab:prompts}
\end{table}

\paragraph{Localization templates}
Entity localization uses templated prompts of the form {\ttfamily The <attribute> of <entity>}. The full template list used in our runs contains 100 attributes, including:
\begin{quote}
\scriptsize
origin, purpose, definition, function, main goal, age, name, founder, owner, value, importance, reputation, impact, influence, location, history, status, category, type, meaning, significance, role, date of creation, latest update, duration, size, popularity, main activity, scope, reach, composition, structure, method, strategy, goal, objective, result, effect, outcome, cause, reason, source, destination, trend, main challenge, opinion, leading opinion, common perception, definition in law, ethical standing, main criticism, key advantage, key disadvantage, limitation, potential, likelihood, probability, risk, opportunity, threat, strength, weakness, main competitor, main supporter, main opponent, relationship with others, relevance, timing, frequency, pattern, cost, budget, revenue, profit, loss, market share, demographic, representation, policy, regulation, requirement, recommendation, limiting factor, resource, technology used, process, legal status, acceptance, approval, recognition, symbolism, associations, link to current events, precedent, measurement, ranking, priority, main feature, unique aspect, distinguishing factor.
\end{quote}

\paragraph{Key hyperparameters} Unless stated otherwise: curated PopQA $N=200$ entities, $K=2$ questions per entity for localization and causal checks, seed 7, and $\epsilon=10^{-6}$ for stability computations. Entity-specific amnesia (main Finding~2) uses $\alpha \in [1, -3]$ with 20 steps.
\paragraph{Finding 3 injection setting} For Finding~3 we use set-injection at the placeholder position with mean-entity initialization. Concretely, we first set the full hidden vector at \texttt{X} to the layer-specific mean-entity vector, then overwrite the selected top-$k$ neurons (single-cell top-1 as primary, with top-5 as the multi-cell comparison). We sweep an interpolation/extrapolation factor $\alpha \in \{1,2,4,8,16,32,64,128,200\}$ per entity, selecting the best-performing $\alpha$ for reporting; this choice is intentionally high-sensitivity and may be optimistic relative to a fixed-$\alpha$ protocol. We additionally flag entity-level success when $\mathrm{RelProb}\ge 0.30$ and both margins $\mathrm{RelProb}-\mathrm{RelProb}_{\mathrm{no\ inj}}\ge 0.05$ and $\mathrm{RelProb}-\mathrm{RelProb}_{\mathrm{wrong}}\ge 0.05$.

\section{Algorithms}
\label{app:algorithms}
Algorithms~\ref{alg:localize} and~\ref{alg:inject} provide pseudocode for the two core procedures used throughout this work: stability-based localization and controlled cell injection. Algorithm~\ref{alg:steer} describes a factual modification procedure via latent steering (Appendix~\ref{app:latent_steering}).

\begin{algorithm}[!ht]
\caption{Stability-based localization of an entity cell}
\label{alg:localize}
\begin{algorithmic}[1]
\Require Model $M$ with layers $\ell \in \{0,\dots,L-1\}$; baseline prompt set $\mathcal{B}$; entity-centered prompts $\{x_i\}_{i=1}^K$; entity token index function $t(\cdot)$; $\epsilon>0$
\Ensure Entity cell $(\ell^\star, j^\star)$
\State Compute baseline statistics $(\mu_{\ell j}, \sigma_{\ell j})$ from $a_{\ell j}(b)$ over $b\in\mathcal{B}$
\For{$i \gets 1$ to $K$}
\State Extract activations $a_{\ell j}(x_i)$ at token position $t(x_i)$ for all $\ell,j$
\State Normalize $z_{\ell j}(x_i) \gets (a_{\ell j}(x_i) - \mu_{\ell j})/(\sigma_{\ell j}+\epsilon)$
\EndFor
\State Compute stability $S_{\ell j} \gets (\mathbb{E}_i[z_{\ell j}(x_i)])^2 / (\mathrm{Std}_i[z_{\ell j}(x_i)] + \epsilon)$
\State $(\ell^\star, j^\star) \gets \arg\max_{\ell,j} S_{\ell j}$
\State \Return $(\ell^\star, j^\star)$
\end{algorithmic}
\end{algorithm}

\begin{algorithm}[!ht]
\caption{Controlled injection of entity cells in a QA-style prompt}
\label{alg:inject}
\begin{algorithmic}[1]
\Require Model $M$; tokenizer $\tau$; PopQA question $q$; answer aliases $\mathcal{A}$; entity aliases $\mathcal{E}$; localized layer $\ell^\star$; top-$k$ entity cells $S=\{j_1,\dots,j_k\}$; entity-specific values $\{v_{j}\}_{j\in S}$; mean-entity vector $m_{\ell^\star}$; scale $\alpha$; $\epsilon>0$
\Ensure Relative answer probability under injection
\State Wrap the question for a base model: $x_{\mathrm{full}} \gets \texttt{Question: }q\texttt{\textbackslash nAnswer:}$
\State Find a matched alias $e \in \mathcal{E}$ in $q$ and form $q_X$ by replacing the first occurrence with \texttt{X}
\State Construct placeholder prompt $x_X \gets \texttt{Question: }q_X\texttt{\textbackslash nAnswer:}$ and locate placeholder token index $t_X$
\State Convert aliases to next-token targets $Y \gets \{\mathrm{tok}_1(a): a\in \mathcal{A}\}$ under $\tau$ (prepend a leading space for tokenization)
\State Run $M$ on $x_{\mathrm{full}}$ and record $p_{\mathrm{full}} \gets \max_{y\in Y} p(y \mid x_{\mathrm{full}})$
\State Run $M$ on $x_X$ (no injection) and record $p_0 \gets \max_{y\in Y} p(y \mid x_X)$
\State Run $M$ on $x_X$ while injecting at layer $\ell^\star$ and position $t_X$:
\Statex \hspace{1em} Initialize: $a_{\ell^\star}(x_X)[t_X] \leftarrow m_{\ell^\star}$
\Statex \hspace{1em} For each $j \in S$: $a_{\ell^\star j}(x_X)[t_X] \leftarrow m_{\ell^\star j} + \alpha\,(v_j - m_{\ell^\star j})$
\State Record $p_1 \gets \max_{y\in Y} p(y \mid x_X,\mathrm{inject})$
\State \Return $p_1 / \max(p_{\mathrm{full}}, \epsilon)$
\end{algorithmic}
\end{algorithm}

\section{Factual Modification via Latent Steering}
\label{app:latent_steering}
We describe a factual modification procedure that optimizes a perturbation vector injected at an entity-associated activation site to increase the probability of a chosen target completion for a specific relation, while penalizing drift on a small set of unrelated facts. We report a single-case study (Obama spouse) to illustrate the method and its edit-vs.-preserve objective.

\begin{table}[!ht]
\centering
\scriptsize
\begingroup
\setlength{\tabcolsep}{2pt}
\renewcommand{\arraystretch}{1.08}
\begin{tabular}{@{}l >{\raggedright\arraybackslash}p{0.82\columnwidth}@{}}
\toprule
\textbf{ID} & \textbf{Prompt (Barack Obama)} \\
\midrule
A1 & The name of the wife of Barack Obama is \\
A2 & When Barack Obama was president, his wife's name was \\
A3 & Barack Obama is married to \\
A4 & The spouse of Barack Obama is named \\
\midrule
P1 & Barack Obama was born in $\rightarrow$ Hawaii \\
P2 & The political party of Barack Obama is $\rightarrow$ Democratic \\
P3 & Barack Obama served as the 44th $\rightarrow$ President \\
P4 & The daughters of Barack Obama are Malia and $\rightarrow$ Sasha \\
P5 & Barack Obama's vice president was Joe $\rightarrow$ Biden \\
P6 & The book written by Barack Obama is titled Dreams from My $\rightarrow$ Father \\
\bottomrule
\end{tabular}
\endgroup
\caption{Attack prompts (A1--A4) and preservation prompts (P1--P6) used for factual modification. Preservation prompts include an expected next token.}
\label{tab:steer_prompts}
\end{table}

\begin{figure}[!ht]
\centering
\includegraphics[width=\linewidth]{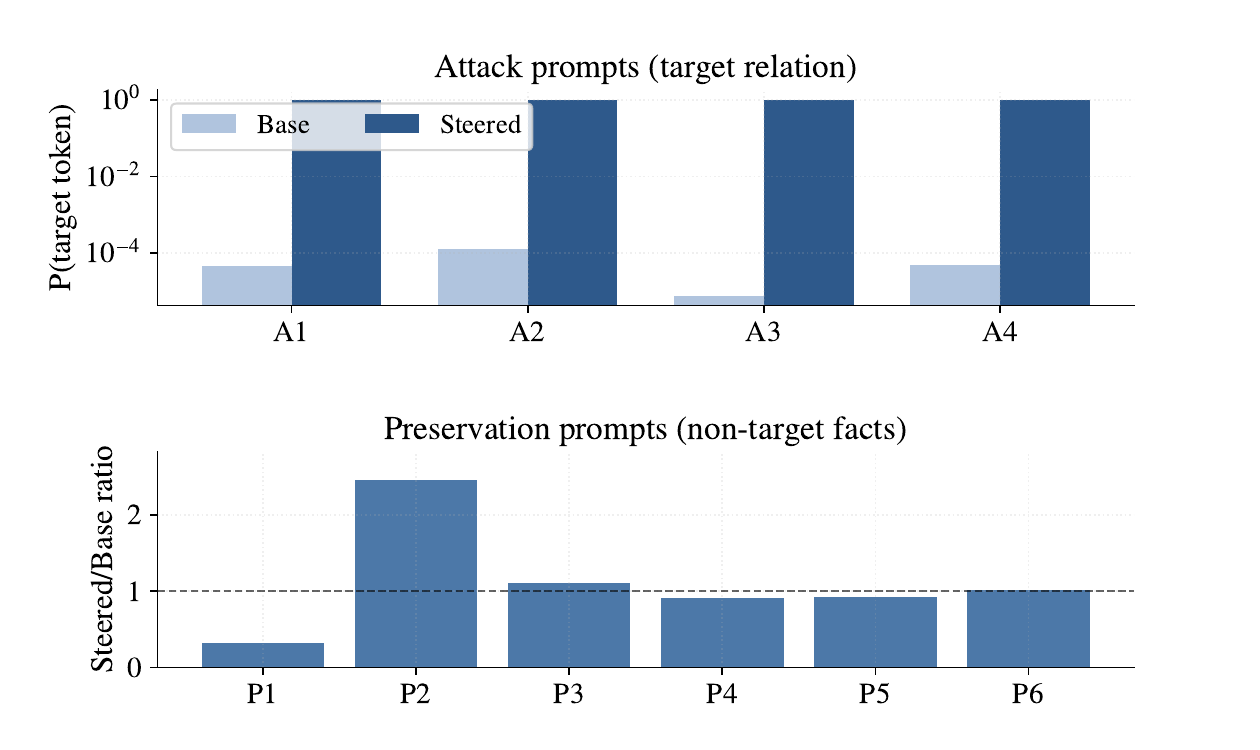}
\caption{Factual modification via latent steering (single-case study). Top: spouse prompts (A1--A4; Table~\ref{tab:steer_prompts}) before/after steering toward a target completion. Bottom: preservation prompts (P1--P6), reported as steered/base ratios for expected next tokens.}
\label{fig:f7}
\end{figure}
\begin{algorithm}[!ht]
\caption{Factual modification via latent steering at a localized entity layer}
\label{alg:steer}
\begin{algorithmic}[1]
\Require Model $M$; tokenizer $\tau$; entity string $e$; localized layer $\ell^\star$; entity token index function $t_e(\cdot)$; attack prompts $\mathcal{A}$; preserve facts $\mathcal{P}=\{(p_i, y_i)\}$; target token $y_{\mathrm{tgt}}$; weights $(\lambda_a,\lambda_p,\lambda_2)$; steps $T$; learning rate $\eta$
\Ensure Perturbation vector $\delta \in \mathbb{R}^{d}$ (hidden size $d$)
\State Initialize $\delta \sim \mathrm{Uniform}(0,1)^d$ (float32), set optimizer $\mathrm{AdamW}(\delta,\eta)$
\For{$t \gets 1$ to $T$}
\State $L_a \gets 0$
\For{each $a \in \mathcal{A}$}
\State Locate entity token index $s \gets t_e(a)$ (e.g., last token of $e$ under $\tau$)
\State Run $M$ on $a$ while injecting $\delta$ at layer $\ell^\star$ and position $s$
\State $L_a \mathrel{+}= -\log p(y_{\mathrm{tgt}} \mid a, \delta)$
\EndFor
\State $L_a \gets \frac{1}{|\mathcal{A}|} L_a$
\State $L_p \gets 0$
\For{each $(p_i,y_i) \in \mathcal{P}$}
\State Locate entity token index $s \gets t_e(p_i)$
\State Run $M$ on $p_i$ while injecting $\delta$ at layer $\ell^\star$ and position $s$
\State $L_p \mathrel{+}= -\log p(y_i \mid p_i, \delta)$
\EndFor
\State $L_p \gets \frac{1}{|\mathcal{P}|} L_p$
\State $L_2 \gets \|\delta\|_2$
\State $\mathcal{L} \gets \lambda_a L_a + \lambda_p L_p + \lambda_2 L_2$
\State Take one optimizer step on $\delta$ using $\nabla_\delta \mathcal{L}$
\EndFor
\State \Return $\delta$
\end{algorithmic}
\end{algorithm}

\newpage
\section{Entity Cell Map}
\label{app:entity_neurons}

\paragraph{Categorized PopQA map}
The full PopQA-200 entity map is grouped by category to improve readability. The current split contains 48 people, 82 locations, 6 organizations, and 64 other entities. Each row includes a trust flag computed by automated checks (e.g., early-layer localization and causal sensitivity under negative ablation, plus non-collapse sanity checks). Under these checks, $k=131$ out of $n=200$ localized cells are marked trustworthy. Under the Finding~3 causal-injection success criterion (full trustworthy set), 75/131 entities pass with top-$k$ injection (74/131 with top-1); 1 entity requires top-$k$.
% Auto-generated from localization + unlearning results with curated category grouping

\begingroup
\setlength{\tabcolsep}{2pt}
\renewcommand{\arraystretch}{1.06}
\scriptsize

\subsection*{Person Entities (k=26, n=48)}
{\centering
\begin{tabular}{@{}>{\raggedright\arraybackslash}p{0.60\columnwidth}rrc@{}}
\toprule
Entity & Layer & Neuron & Trust. \\
\midrule
Abraham Lincoln & 3 & 12305 & Yes \\
Al Gore & 1 & 11620 & Yes \\
Alexander the Great & 4 & 8881 & Yes \\
Ali ibn Abi Talib & 4 & 18599 & No \\
Amitabh Bachchan & 3 & 10957 & No \\
Apollo & 0 & 8859 & No \\
Aung San Suu Kyi & 3 & 3083 & No \\
Barack Obama & 2 & 10941 & Yes \\
Bertrand Russell & 3 & 15419 & Yes \\
Billy Joel & 2 & 8277 & Yes \\
Carl Linnaeus & 2 & 18724 & No \\
Chris Jericho & 3 & 8819 & Yes \\
David & 2 & 13244 & Yes \\
Donald Trump & 1 & 11948 & Yes \\
Edgar Allan Poe & 5 & 16637 & Yes \\
Elizabeth II & 3 & 6343 & No \\
Francis & 2 & 2926 & Yes \\
Ganesha & 2 & 5215 & No \\
\bottomrule
\end{tabular}\par}
\smallskip

{\centering
\begin{tabular}{@{}>{\raggedright\arraybackslash}p{0.60\columnwidth}rrc@{}}
\toprule
Entity & Layer & Neuron & Trust. \\
\midrule
Gautama Buddha & 0 & 14566 & No \\
George VI & 1 & 467 & Yes \\
George W. Bush & 0 & 10032 & No \\
George Washington & 1 & 3732 & No \\
Hamilton & 4 & 15761 & No \\
Helen of Troy & 2 & 17460 & No \\
Jacob & 1 & 4 & Yes \\
James Madison & 2 & 3867 & No \\
James VI and I & 0 & 698 & No \\
Jesus & 1 & 8526 & Yes \\
Johann Sebastian Bach & 4 & 11153 & No \\
King Arthur & 4 & 6331 & No \\
Krishna & 3 & 12927 & Yes \\
Mark Twain & 1 & 11338 & No \\
Mary, Princess Royal and Countess of Harewood & 0 & 12945 & No \\
Michael Jackson & 2 & 1224 & Yes \\
Muhammad & 2 & 18938 & No \\
Muhammad Ali & 1 & 17041 & No \\
\bottomrule
\end{tabular}\par}
\smallskip

{\centering
\begin{tabular}{@{}>{\raggedright\arraybackslash}p{0.60\columnwidth}rrc@{}}
\toprule
Entity & Layer & Neuron & Trust. \\
\midrule
Paul & 1 & 18738 & Yes \\
Peter & 2 & 13512 & Yes \\
Prince & 1 & 15522 & Yes \\
Queen Victoria & 1 & 18579 & Yes \\
Rama & 4 & 18757 & Yes \\
Ronan Farrow & 2 & 5954 & No \\
Rumi & 2 & 6292 & Yes \\
T. S. Eliot & 1 & 12920 & Yes \\
Thomas Jefferson & 2 & 11279 & Yes \\
Thor & 4 & 6521 & Yes \\
Vajiralongkorn & 2 & 3566 & No \\
Will Smith & 3 & 15898 & Yes \\
\bottomrule
\end{tabular}\par}
\smallskip

\medskip
\subsection*{Location Entities (k=56, n=82)}
{\centering
\begin{tabular}{@{}>{\raggedright\arraybackslash}p{0.60\columnwidth}rrc@{}}
\toprule
Entity & Layer & Neuron & Trust. \\
\midrule
Afghanistan & 3 & 2492 & Yes \\
Alexandria & 2 & 9910 & No \\
Arizona & 3 & 8202 & No \\
Arkansas & 3 & 3613 & No \\
Athens & 1 & 18605 & Yes \\
Australia & 1 & 8982 & Yes \\
Barcelona & 3 & 13144 & Yes \\
Beijing & 2 & 9567 & Yes \\
Berlin & 2 & 17703 & No \\
Boston & 3 & 18601 & Yes \\
Brazil & 1 & 17255 & Yes \\
Brussels Capital Region & 0 & 7084 & No \\
Byzantine Empire & 2 & 1260 & Yes \\
California & 2 & 4130 & No \\
Canada & 2 & 15999 & Yes \\
Cape Town & 3 & 9847 & Yes \\
Chicago & 3 & 10161 & No \\
China & 2 & 6806 & No \\
\bottomrule
\end{tabular}\par}
\smallskip

{\centering
\begin{tabular}{@{}>{\raggedright\arraybackslash}p{0.60\columnwidth}rrc@{}}
\toprule
Entity & Layer & Neuron & Trust. \\
\midrule
Colorado & 2 & 12787 & No \\
Confederate States of America & 0 & 8161 & No \\
Dallas & 3 & 3153 & Yes \\
Delhi & 4 & 7686 & Yes \\
Dublin & 4 & 3121 & Yes \\
El Salvador & 1 & 2281 & No \\
Empire of Japan & 1 & 7726 & Yes \\
Florence & 1 & 10577 & Yes \\
Florida & 2 & 7558 & No \\
Georgia & 1 & 12233 & Yes \\
Hawaii & 1 & 12270 & Yes \\
Houston & 1 & 11316 & Yes \\
Idaho & 4 & 649 & Yes \\
India & 2 & 2893 & Yes \\
Italy & 1 & 2937 & No \\
Japan & 1 & 7726 & Yes \\
Jerusalem & 3 & 270 & Yes \\
Jordan & 2 & 1661 & No \\
\bottomrule
\end{tabular}\par}
\smallskip

{\centering
\begin{tabular}{@{}>{\raggedright\arraybackslash}p{0.60\columnwidth}rrc@{}}
\toprule
Entity & Layer & Neuron & Trust. \\
\midrule
Kansas & 2 & 11619 & Yes \\
Kuala Lumpur & 3 & 2420 & Yes \\
Lebanon & 1 & 4918 & No \\
London & 2 & 17407 & Yes \\
Madrid & 2 & 10473 & Yes \\
Mali & 3 & 10631 & Yes \\
Manila & 3 & 16721 & No \\
Melbourne & 1 & 5250 & Yes \\
Mexico & 1 & 977 & No \\
Milan & 2 & 13790 & No \\
Minnesota & 2 & 5780 & Yes \\
Montana & 2 & 13501 & Yes \\
Nebraska & 2 & 17105 & Yes \\
Netherlands & 3 & 15677 & No \\
New Jersey & 2 & 10170 & Yes \\
New Mexico & 4 & 10866 & Yes \\
New York & 1 & 11260 & No \\
New York City & 1 & 1139 & Yes \\
\bottomrule
\end{tabular}\par}
\smallskip

{\centering
\begin{tabular}{@{}>{\raggedright\arraybackslash}p{0.60\columnwidth}rrc@{}}
\toprule
Entity & Layer & Neuron & Trust. \\
\midrule
Oregon & 4 & 4925 & Yes \\
Paris & 1 & 231 & Yes \\
Peru & 2 & 17476 & Yes \\
Philadelphia & 4 & 751 & Yes \\
Phoenix & 1 & 11122 & No \\
Pittsburgh & 1 & 8853 & Yes \\
Poland & 4 & 7987 & Yes \\
Prague & 3 & 14586 & Yes \\
Puerto Rico & 2 & 7844 & Yes \\
Republic of China 1912-1949 & 0 & 6408 & No \\
Rio de Janeiro & 1 & 1640 & No \\
Roman Republic & 1 & 6083 & No \\
Rome & 4 & 5848 & Yes \\
San Francisco & 1 & 9914 & Yes \\
Singapore & 1 & 7100 & Yes \\
South Africa & 1 & 1385 & Yes \\
Spain & 2 & 5876 & Yes \\
Sri Lanka & 2 & 6010 & Yes \\
\bottomrule
\end{tabular}\par}
\smallskip

{\centering
\begin{tabular}{@{}>{\raggedright\arraybackslash}p{0.60\columnwidth}rrc@{}}
\toprule
Entity & Layer & Neuron & Trust. \\
\midrule
Stockholm & 5 & 14077 & Yes \\
Tennessee & 3 & 16787 & Yes \\
Texas & 3 & 4501 & Yes \\
Tokyo & 1 & 188 & Yes \\
Toronto & 1 & 864 & Yes \\
Troy & 2 & 17460 & Yes \\
Turin & 3 & 18612 & Yes \\
Vienna & 1 & 18529 & Yes \\
Virginia & 1 & 7229 & No \\
Washington, D.C. & 0 & 16991 & No \\
\bottomrule
\end{tabular}\par}
\smallskip

\medskip
\subsection*{Organization Entities (k=4, n=6)}
{\centering
\begin{tabular}{@{}>{\raggedright\arraybackslash}p{0.60\columnwidth}rrc@{}}
\toprule
Entity & Layer & Neuron & Trust. \\
\midrule
Atlético de Madrid & 2 & 10473 & Yes \\
European Union & 2 & 12264 & Yes \\
Nine Inch Nails & 3 & 5818 & No \\
Oasis & 3 & 13059 & Yes \\
The Band & 0 & 8075 & No \\
White House & 1 & 18670 & Yes \\
\bottomrule
\end{tabular}\par}
\smallskip

\medskip
\subsection*{Other Entities (k=45, n=64)}
{\centering
\begin{tabular}{@{}>{\raggedright\arraybackslash}p{0.60\columnwidth}rrc@{}}
\toprule
Entity & Layer & Neuron & Trust. \\
\midrule
19 & 0 & 13033 & No \\
Alien & 3 & 10404 & Yes \\
Aliens & 1 & 7483 & Yes \\
Avatar & 2 & 14834 & No \\
Babylon & 3 & 11790 & Yes \\
Back to the Future & 0 & 11486 & No \\
Battlefield & 2 & 10945 & Yes \\
Beloved & 1 & 7245 & Yes \\
Breaking Bad & 2 & 12152 & Yes \\
Budapest & 3 & 10222 & Yes \\
Carrie & 3 & 18279 & Yes \\
Cars & 1 & 18321 & Yes \\
Doctor Who & 0 & 1329 & No \\
Drive & 2 & 16166 & No \\
E.T. the Extra-Terrestrial & 1 & 4102 & Yes \\
Final Destination & 2 & 12738 & Yes \\
Flight & 2 & 2518 & Yes \\
Friends & 5 & 9314 & Yes \\
\bottomrule
\end{tabular}\par}
\smallskip

{\centering
\begin{tabular}{@{}>{\raggedright\arraybackslash}p{0.60\columnwidth}rrc@{}}
\toprule
Entity & Layer & Neuron & Trust. \\
\midrule
Frozen & 2 & 16115 & Yes \\
Ghost & 1 & 4299 & No \\
Grease & 2 & 10749 & Yes \\
Halloween & 4 & 2177 & No \\
Happy Birthday to You & 5 & 2162 & Yes \\
Heart & 1 & 17053 & No \\
Inside Out & 0 & 2940 & No \\
Into the Wild & 1 & 8382 & Yes \\
Iron Man & 1 & 16367 & No \\
It & 0 & 10424 & No \\
Jesus in Islam & 1 & 5588 & Yes \\
Legend & 0 & 5788 & No \\
Léon: The Professional & 2 & 9734 & Yes \\
Let It Be & 5 & 11669 & Yes \\
Lost & 1 & 5217 & Yes \\
Neon Genesis Evangelion & 4 & 4605 & Yes \\
Nineteen Eighty-Four & 1 & 12211 & No \\
Power & 22 & 12984 & No \\
\bottomrule
\end{tabular}\par}
\smallskip

{\centering
\begin{tabular}{@{}>{\raggedright\arraybackslash}p{0.60\columnwidth}rrc@{}}
\toprule
Entity & Layer & Neuron & Trust. \\
\midrule
Rent & 2 & 11038 & Yes \\
Rocky & 3 & 6169 & Yes \\
Saw & 27 & 12646 & No \\
Scooby-Doo & 2 & 18490 & Yes \\
Seven & 2 & 9649 & Yes \\
Star Wars & 1 & 6101 & Yes \\
Suits & 2 & 9620 & Yes \\
The Birds & 2 & 8054 & Yes \\
The Challenge & 1 & 13535 & No \\
The Departed & 5 & 2625 & Yes \\
The Fly & 2 & 10777 & Yes \\
The Fog & 1 & 10964 & Yes \\
The Graduate & 2 & 17192 & No \\
The Holiday & 1 & 2834 & Yes \\
The Matrix & 1 & 613 & No \\
The Mist & 2 & 13859 & Yes \\
The Omen & 3 & 5504 & Yes \\
The Prestige & 1 & 1538 & Yes \\
\bottomrule
\end{tabular}\par}
\smallskip

{\centering
\begin{tabular}{@{}>{\raggedright\arraybackslash}p{0.60\columnwidth}rrc@{}}
\toprule
Entity & Layer & Neuron & Trust. \\
\midrule
The Ring & 2 & 14458 & Yes \\
The Shining & 5 & 9949 & Yes \\
The Social Network & 2 & 804 & No \\
The Terminal & 2 & 16927 & Yes \\
The Thing & 5 & 10410 & Yes \\
The Village & 1 & 14056 & Yes \\
They Live & 3 & 18417 & Yes \\
Toy Story & 3 & 2434 & Yes \\
Training Day & 4 & 7991 & Yes \\
Up & 2 & 14580 & Yes \\
\bottomrule
\end{tabular}\par}
\smallskip

\medskip
\endgroup

%\clearpage
\section{Post-Training Generalization (Qwen2.5-7B-Instruct)}
\label{app:qwen25_family}
We test whether the Qwen2.5 entity-cell map survives ordinary post-training by rerunning the same analyses on Qwen2.5-7B-Instruct and comparing directly to the base-model findings in \Cref{fig:layerhist,fig:f6,fig:f4,fig:f3,fig:f3_acronym,fig:f3_multilingual}. Across the full PopQA-200 inventory, Qwen2.5-7B-Instruct exactly preserves the base model's top localized cell for 190/200 entities (and preserves the same layer for 191/200). In particular, both models localize Barack Obama to the same cell (L2-N10941).

This is the clearest post-training result in the appendix. The localization map remains nearly unchanged, the early-layer concentration pattern is preserved, and 123 cells still pass the same amnesia-based trust filter used in the main paper. Together, these results suggest that the Qwen2.5 entity-cell map is robust to ordinary instruction tuning.

\begin{figure}[!ht]
\centering
\includegraphics[width=\linewidth]{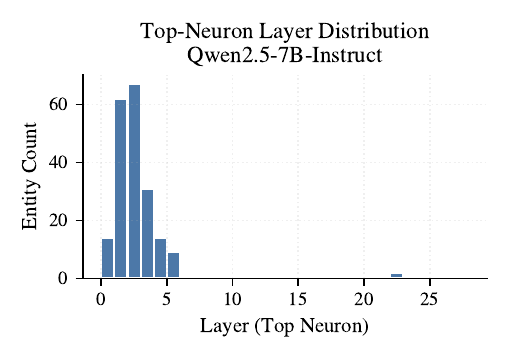}
\caption{Qwen2.5-7B-Instruct replication of \Cref{fig:layerhist}. As in the main-paper Qwen2.5-7B base result, top localized cells remain concentrated in early layers, indicating that the layer profile is largely preserved under instruction tuning.}
\label{fig:qwen25inst_f2}
\end{figure}

\begin{figure}[!ht]
\centering
\includegraphics[width=\linewidth]{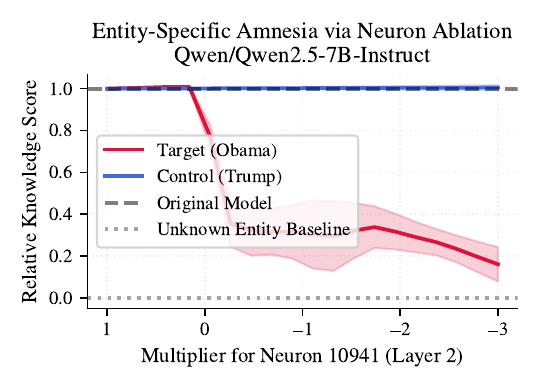}
\caption{Qwen2.5-7B-Instruct replication of \Cref{fig:f6}. Negative ablation of the localized entity cell again causes a strong drop for the target entity while leaving the control entity comparatively stable, supporting preservation of the same causal pattern after post-training.}
\label{fig:qwen25inst_f6}
\end{figure}

\begin{figure}[!ht]
\centering
\includegraphics[width=\linewidth]{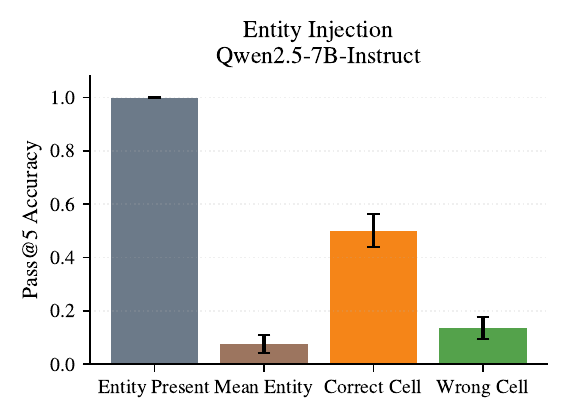}
\caption{Qwen2.5-7B-Instruct replication of \Cref{fig:f4}. On the trusted set, correct-cell injection again outperforms both the mean-entity initialization and wrong-cell controls, indicating that the same localized cells remain causally useful after instruction tuning.}
\label{fig:qwen25inst_f4}
\end{figure}

\begin{figure}[!ht]
\centering
\makebox[\linewidth][c]{\includegraphics[width=1.14\linewidth]{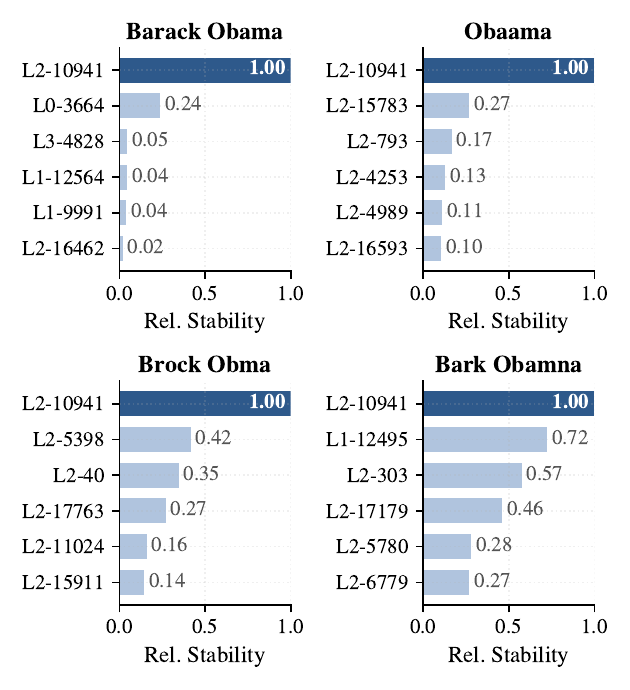}}
\caption{Qwen2.5-7B-Instruct replication of \Cref{fig:f3}. Most variants of ``Barack Obama'' preserve the same top cell as in the base model, indicating that the localized handle remains stable under post-training.}
\label{fig:qwen25inst_f3_variants}
\end{figure}

\begin{figure}[!ht]
\centering
\makebox[\linewidth][c]{\includegraphics[width=1.14\linewidth]{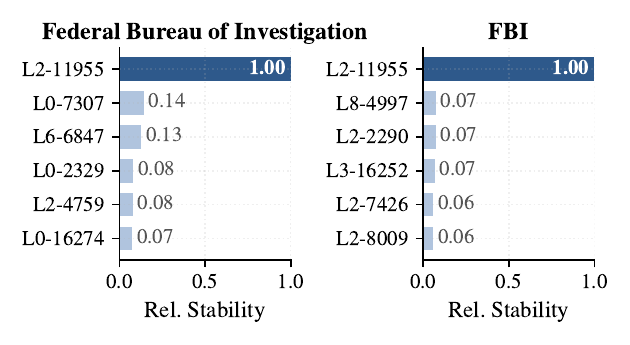}}
\caption{Qwen2.5-7B-Instruct replication of \Cref{fig:f3_acronym}. The acronym and full form again localize to the same top-ranked cell, consistent with preservation of the underlying entity-cell mapping.}
\label{fig:qwen25inst_f3_acronym}
\end{figure}

\begin{figure}[!ht]
\centering
\makebox[\linewidth][c]{\includegraphics[width=1.14\linewidth]{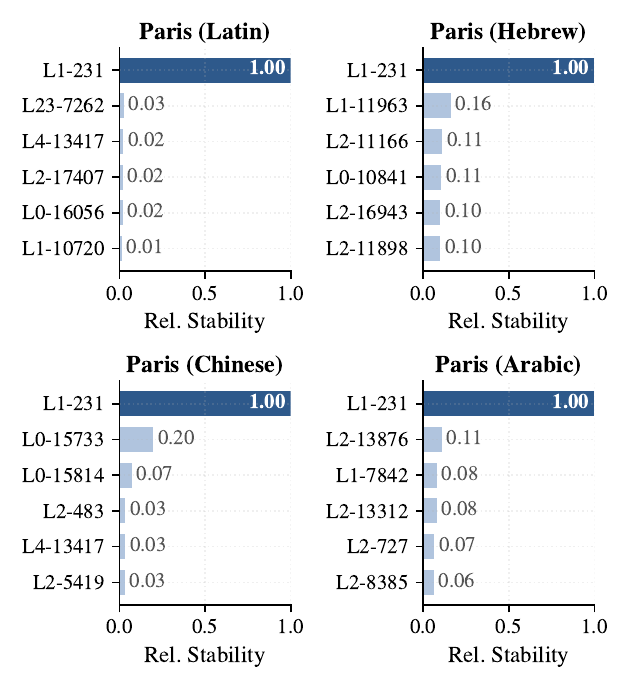}}
\caption{Qwen2.5-7B-Instruct replication of \Cref{fig:f3_multilingual}. The same top cell is recovered across multiple scripts for ``Paris'', suggesting that cross-script entity access also survives instruction tuning.}
\label{fig:qwen25inst_f3_multilingual}
\end{figure}

\newpage
\section{Generalization Within Model Family (Qwen3)}
\label{app:qwen3}
To probe within-family generalization beyond post-training, we apply the same analyses to Qwen3-8B base and report the same figure set used for Qwen2.5-7B-Instruct, matched to the main-paper results in \Cref{fig:layerhist,fig:f6,fig:f4,fig:f3,fig:f3_acronym,fig:f3_multilingual}. The depth profile remains early-layer concentrated, and the larger appendix suite yields stable top localized cells for all 200 PopQA-200 entities. Under the strict trustworthy plus entity-pass@5 filter used for the controlled-injection evaluation, 42 entities remain.

The result is mixed but still suggestive of within-family continuity. Qwen3 reproduces the early-layer localization pattern and retains a nontrivial trustworthy subset, but the causal evidence is weaker than in Qwen2.5. Because the standard Obama/Trump amnesia probe is noisier in Qwen3, we also report an alternative London/Paris unlearning probe using the localized London cell (L0-N3037), which yields cleaner target-versus-control separation. Table~\ref{tab:qwen3_cells} lists representative localized cells.

\begin{figure}[!ht]
\centering
\includegraphics[width=\linewidth]{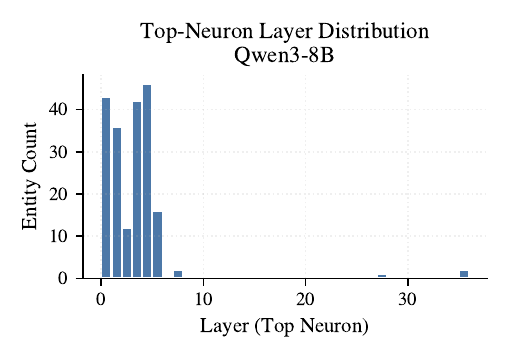}
\caption{Qwen3-8B within-family replication of \Cref{fig:layerhist}. As in the main-paper Qwen2.5 result, top localized cells remain concentrated in early layers, suggesting that the coarse localization pattern persists within the Qwen family.}
\label{fig:qwen3_f2}
\end{figure}

\begin{figure}[!ht]
\centering
\includegraphics[width=\linewidth]{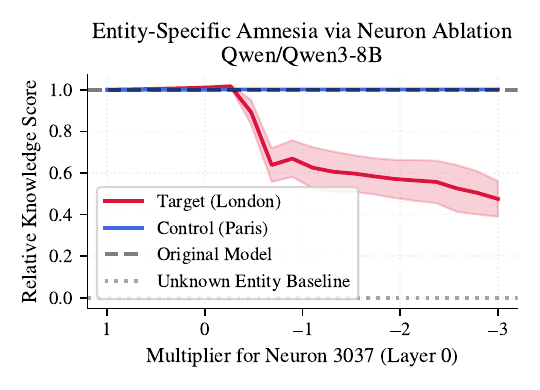}
\caption{Qwen3-8B within-family replication of \Cref{fig:f6}, shown on an alternative London/Paris probe because it yields cleaner separation than the default Obama/Trump case in this model. Negative ablation still preferentially suppresses the target-entity curve, but the effect is less clean than in Qwen2.5.}
\label{fig:qwen3_f6}
\end{figure}

\begin{figure}[!ht]
\centering
\includegraphics[width=\linewidth]{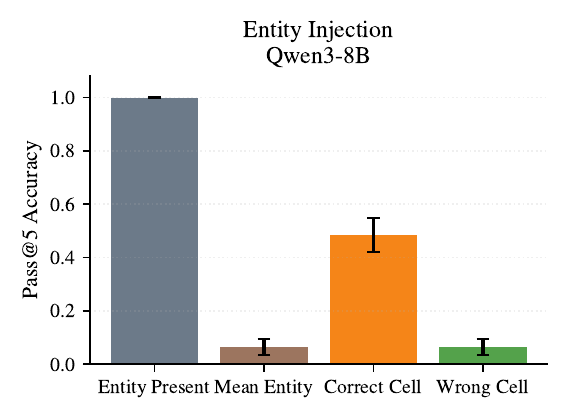}
\caption{Qwen3-8B within-family replication of \Cref{fig:f4}. Correct-cell injection improves over the control conditions, but the separation is weaker and less consistent than in Qwen2.5, matching the more mixed within-family result described in the text.}
\label{fig:qwen3_f4}
\end{figure}

\begin{figure}[!ht]
\centering
\makebox[\linewidth][c]{\includegraphics[width=1.14\linewidth]{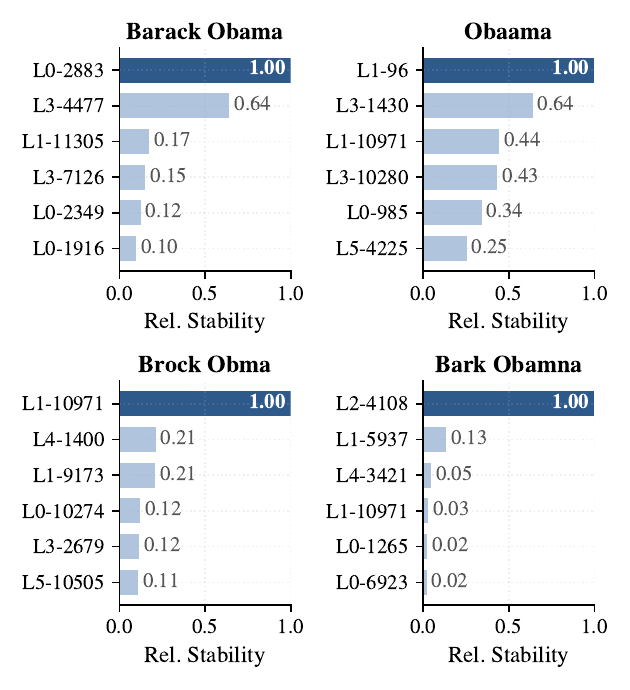}}
\caption{Qwen3-8B within-family replication of \Cref{fig:f3}. Most variants recover closely related early-layer cells, though the match is less clean than in Qwen2.5-7B-Instruct.}
\label{fig:qwen3_f3_variants}
\end{figure}

\begin{figure}[!ht]
\centering
\makebox[\linewidth][c]{\includegraphics[width=1.14\linewidth]{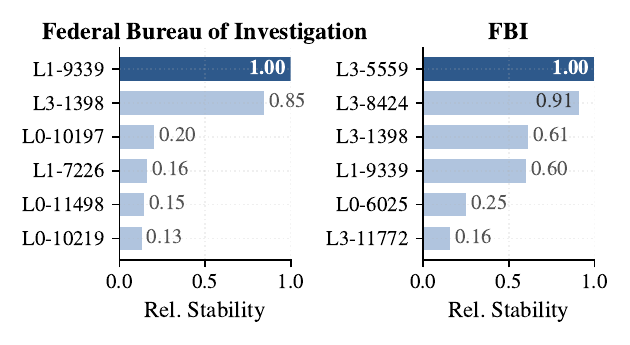}}
\caption{Qwen3-8B within-family replication of \Cref{fig:f3_acronym}. The acronym and expanded form still localize to closely aligned cells, supporting partial preservation of the entity-cell map within the Qwen family.}
\label{fig:qwen3_f3_acronym}
\end{figure}

\begin{figure}[!ht]
\centering
\makebox[\linewidth][c]{\includegraphics[width=1.14\linewidth]{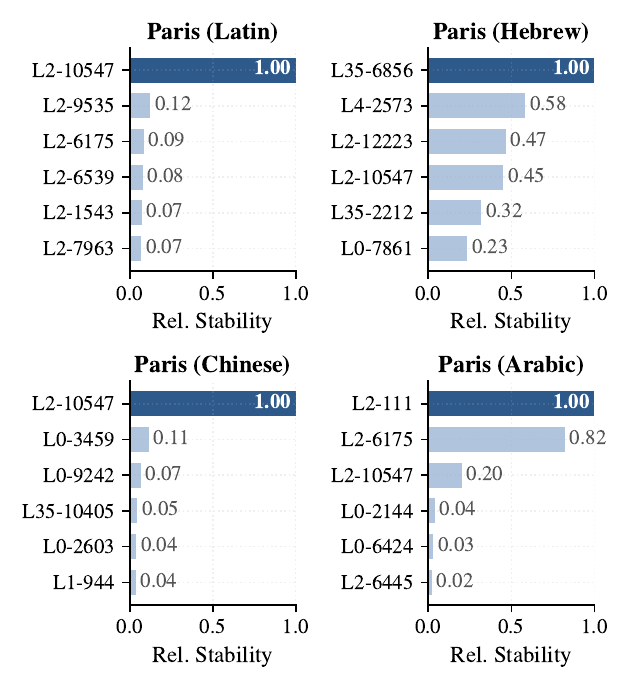}}
\caption{Qwen3-8B within-family replication of \Cref{fig:f3_multilingual}. Cross-script forms continue to recover similar top cells, though the pattern is noisier than in Qwen2.5.}
\label{fig:qwen3_f3_multilingual}
\end{figure}

\begin{table}[!ht]
\centering
\scriptsize
\begingroup
\setlength{\tabcolsep}{2pt}
\renewcommand{\arraystretch}{1.08}
\begin{tabular}{@{}
  >{\raggedright\arraybackslash}p{0.27\columnwidth}
  rr
  >{\raggedright\arraybackslash}p{0.35\columnwidth}
@{}}
\toprule
\textbf{Entity} & \textbf{Layer} & \textbf{Neuron} & \textbf{Notes} \\
\midrule
Obama & 0 & 2883 & localized (injection weak in this subset) \\
Trump & 3 & 9290 & localized; injection success \\
Paris & 2 & 10547 & localized \\
London & 0 & 3037 & localized; strong ablation drop \\
Beijing & 4 & 5431 & localized; strong ablation drop \\
Tokyo & 3 & 223 & localized; ablation drop \\
\bottomrule
\end{tabular}
\endgroup
\caption{Representative Qwen3-8B base entity cells localized by stability. Notes summarize small-subset causal checks (Appendix~\ref{app:qwen3}).}
\label{tab:qwen3_cells}
\end{table}

\section{Lack of Generalization Across Model Families}
\label{app:cross_model}
To summarize cross-family transfer, we use a simple count-based pipeline for each model: start from the same 200 PopQA entities, localize one top cell per entity, keep only entities that pass the Finding~2 amnesia trust filter, evaluate controlled injection on this trusted set, and test exact top-cell stability across surface-form probes.

Table~\ref{tab:cross_model_summary} shows that cross-family transfer is limited. OLMo-7B gives the strongest result: 37 trustworthy cells, 23 of which pass controlled injection, with 30\% form robustness. The other families are weaker. Llama-3.1-8B and Mistral-7B each retain 40 trustworthy cells, but only 5 pass injection; OpenLLaMA-7B retains 33 trustworthy cells, of which 12 pass injection. Overall, sparse candidate cells can often be localized, but strong causal validation and form robustness do not transfer reliably.

\begin{figure}[!ht]
\centering
\includegraphics[width=\linewidth]{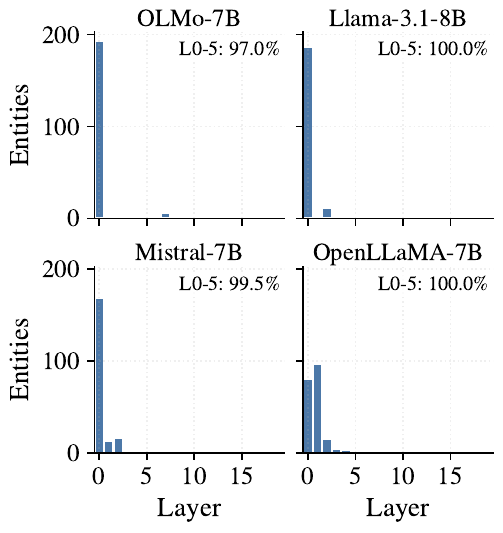}
\caption{Cross-model localization depth profiles (2$\times$2) for four non-Qwen-family models, formatted for single-column display. Each panel shows the distribution of top-neuron layers over PopQA-200 entities for one model. Relative to the main-paper localization result in \Cref{fig:layerhist}, these families are generally broader and deeper.}
\label{fig:cross_model_f2_grid}
\end{figure}

\Cref{fig:cross_model_f2_grid} provides a localization-only comparison across four non-Qwen model families. Relative to the dedicated Qwen-family plots and to the main-paper localization result in \Cref{fig:layerhist}, these distributions are typically broader and shifted deeper, suggesting that the sparse early-layer localization pattern is not uniform across model families.

\begin{figure}[!ht]
\centering
\includegraphics[width=\linewidth]{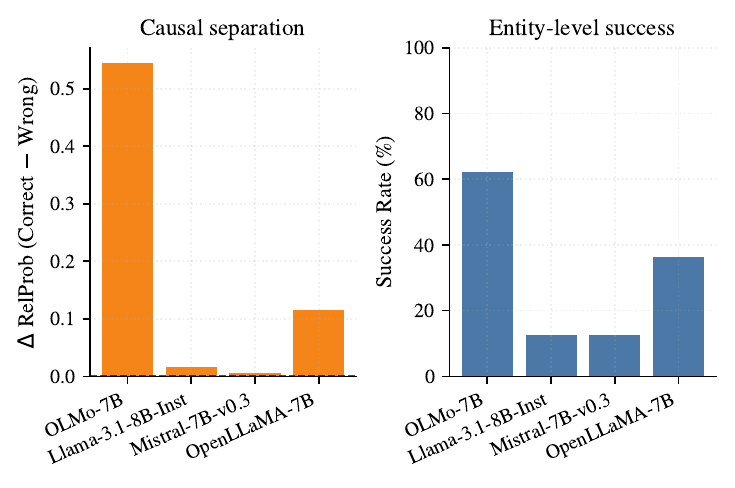}
\caption{Controlled injection on trustworthy cells (top-$5$, alpha search). Left: causal separation $\Delta=\mathrm{RelProb}(\text{Correct Cell})-\mathrm{RelProb}(\text{Wrong Cell})$, where probabilities are normalized by the entity-present prompt. Right: entity-level success rate, defined as the fraction of trustworthy entities that satisfy all three criteria: $\mathrm{RelProb}(\text{Correct Cell})\ge 0.30$, improvement over no-injection $\ge 0.05$, and improvement over wrong-cell injection $\ge 0.05$.}
\label{fig:cross_model_f4}
\end{figure}

\Cref{fig:cross_model_f4} summarizes controlled injection on the trusted set for each model. As in the main-paper injection analysis in \Cref{fig:f4}, we replace the entity mention with a placeholder token, inject either the matched localized cell(s) or a wrong entity's cell(s) at that position, and compare normalized answer probability to the entity-present baseline.

We include a dedicated OLMo view because it shows the strongest cross-family controlled injection result in Table~\ref{tab:cross_model_summary} ($23/37$). At the same time, its amnesia curve is less clean than the Qwen-family cases, suggesting that these cells may play a broader or differently structured role in OLMo.

\begin{figure}[!ht]
\centering
\includegraphics[width=0.94\linewidth]{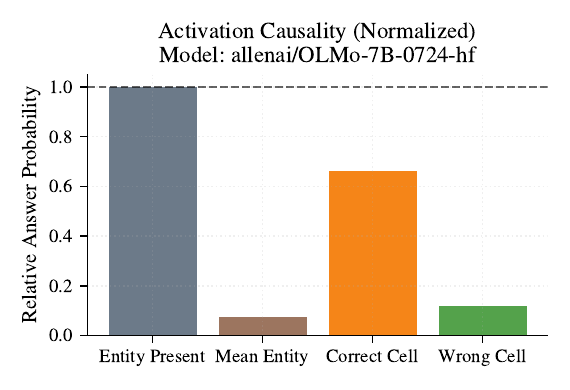}
\caption{OLMo-7B cross-family replication of \Cref{fig:f4}. Among the non-Qwen models, OLMo shows the strongest positive result: correct-cell injection produces the clearest separation from the control conditions on the trusted set.}
\label{fig:olmo_f4}
\end{figure}

\begin{figure}[!ht]
\centering
\includegraphics[width=\linewidth]{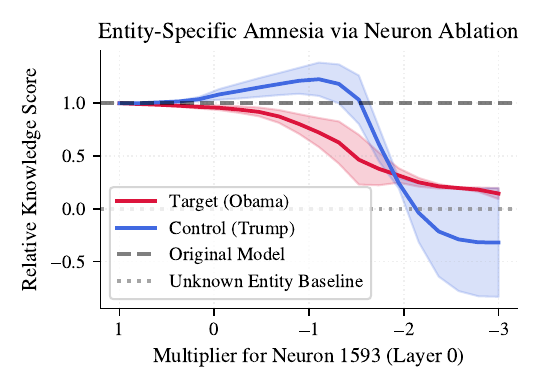}
\caption{OLMo-7B cross-family replication of \Cref{fig:f6}. Negative ablation reduces the target-entity curve, but the control curve is also affected, so this is not the same clean entity-specific amnesia pattern seen in Qwen2.5. This suggests that OLMo's localized cells may participate in retrieval differently, or less selectively, than the Qwen-family cells.}
\label{fig:olmo_f6}
\end{figure}

\begin{table}[!ht]
\centering
\footnotesize
\setlength{\tabcolsep}{1.5pt}
\begin{tabular}{@{}lccc@{}}
\toprule
\textbf{Model} & \textbf{Trustworthy} & \textbf{Injection} & \textbf{ Form-Robustness} \\
\midrule
OLMo-7B & 37 & 23/37 & 30\% \\
Llama-3.1-8B & 40 & 5/40 & 50\% \\
Mistral-7B & 40 & 5/40 & 40\% \\
OpenLLaMA-7B & 33 & 12/33 & 30\% \\
\bottomrule
\end{tabular}%
\caption{Cross-family summary over PopQA-200 entities. \emph{Trustworthy Cells}: entities retained by the Finding~2 amnesia trust filter. \emph{Knowledge Injection}: top-$5$ success among trustworthy entities. \emph{Surface-Form Robustness}: exact top-cell match rate across variant, acronym, and multilingual probes (10 attempts per probe).}
\label{tab:cross_model_summary}
\end{table}
\end{document}